\pdfoutput=1
\documentclass[journal]{IEEEtran}
\usepackage{cite}
\usepackage{amsmath,amssymb,amsfonts}
\usepackage{graphicx}
\usepackage{booktabs}
\usepackage{array}
\usepackage{textcomp}
\usepackage{xcolor}
\usepackage{url}
\usepackage[hidelinks]{hyperref}
\usepackage[all]{hypcap}
\DeclareRobustCommand{\figref}[2][]{\hyperref[#2]{Fig.~\ref*{#2}#1}}
\DeclareRobustCommand{\tabref}[1]{\hyperref[#1]{\mbox{Table}~\ref*{#1}}}
\DeclareRobustCommand{\secref}[1]{\hyperref[#1]{\mbox{Section}~\ref*{#1}}}
\graphicspath{{figures/}}
\newcommand{\tb}[1]{\textbf{#1}}
\begin{document}
\bstctlcite{IEEEexample:BSTcontrol}
\title{HierarchiRetina: A Diabetic Retinopathy Screening and Grading Cascade with Heterogeneous Mixture-of-Experts Lesion Segmentation}

\author{Md Asif Uddin,
        Niloy Farhan,
        Rafeed Rahman,
        and~Md. Golam Rabiul Alam%
\thanks{This work has been submitted to the IEEE for possible publication. Copyright may be transferred without notice, after which this version may no longer be accessible.}%
\thanks{This work received no external funding. \emph{(Corresponding author: Md Asif Uddin.)}}%
\thanks{The authors are with the Department of Computer Science and Engineering, Brac University, Dhaka, Bangladesh (e-mail: md.asif.uddin@g.bracu.ac.bd;\\ niloy.farhan@bracu.ac.bd; rahman.rafeed@bracu.ac.bd;\\ rabiul.alam@bracu.ac.bd).}}

\markboth{IEEE Journal of Biomedical and Health Informatics}%
{Uddin \MakeLowercase{\textit{et al.}}: HierarchiRetina}

\maketitle

\begin{abstract}
Clinical screening for diabetic retinopathy proceeds in stages, yet most automated systems grade fundus images with a single black-box network. Three gaps remain: (i) lesions differ widely in size and contrast, but segmenters typically apply the same blocks to all; (ii) ungradable images are usually excluded or treated as an extra class, and lesion-aware graders seldom test whether lesions drive the grade; (iii) staged systems often report each stage only on the images reaching it, leaving propagated errors unmeasured. HierarchiRetina, a screening, segmentation and grading cascade with inspectable stage outputs, addresses each gap. First, HSMoE-AUNet is a U-Net whose skip, bottleneck and decoder blocks are sparse mixtures of \emph{heterogeneous} experts, with lesion-matched types, counts and losses; SwinHRUNetPP, a dual-encoder vessel segmenter, uses a field-of-view-restricted loss. Second, LG-DRG separates gradability from ordinal severity and lets lesion masks query the image through gated cross-attention. Third, an end-to-end evaluation counts screening misses as errors and traces each error to its stage. Pooled Dice ranged from 0.184 (microaneurysms) to 0.579 (exudates); vessel Dice was 0.731. On 58,689 held-out images from six datasets, end-to-end quadratic weighted kappa was 0.804, with referral sensitivity 0.832 at specificity 0.947, and the screening gate caused 66\% of errors. In ablations, the grade ignored the masks, even with the cross-attention gate started open. On all 4,105 official DDR test images, six-class accuracy was 85.6\%, above the best published six-class result we found (83.8\%). Staged screening systems should be evaluated end-to-end, with stage attribution and mask ablation, before clinical deployment.
\end{abstract}


\section{Introduction}
\IEEEPARstart{D}{iabetic} retinopathy (DR) is a leading cause of preventable vision loss in working-age adults~\cite{teo2021global}. Fundus photographs are graded on a five-level scale~\cite{wilkinson2003proposed}, and some cannot be graded at all. Human graders typically decide first whether any retinopathy is present, then look for the lesions that define the grade (microaneurysms, MA; haemorrhages, HE; hard exudates, EX; cotton-wool spots, CWS), and assign a severity level or refer an image that cannot be read. Deep networks reach specialist-level referral accuracy~\cite{gulshan2016development}, but most are black boxes that map the image to a grade.

\begin{figure*}[!t]
\centering
\includegraphics[width=\textwidth]{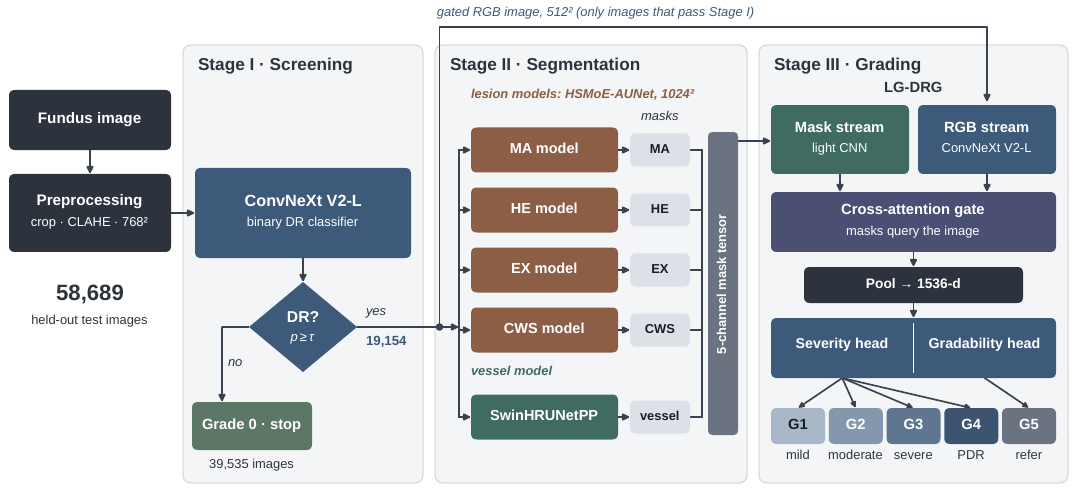}
\caption{HierarchiRetina. Stage~I stops 39,535 test images as Grade~0; Stage~II segments the 19,154 images that pass, using four HSMoE-AUNet lesion models and SwinHRUNetPP; Stage~III (LG-DRG) outputs Grades~1 to 4 or Grade~5 (refer). CLAHE, contrast-limited adaptive histogram equalisation; PDR, proliferative DR.}
\label{fig:overview}
\end{figure*}

\subsection{Related Work}
Recent graders report gains on DDR~\cite{li2019ddr} with ordinal transformers~\cite{yu2026dualswinord}, Radon-transform fusion~\cite{mohsen2025radfuse} and knowledge-guided attention~\cite{tian2023fakcnet}, but their splits, class definitions and handling of ungradable images differ, which hampers comparison. Lesions have entered grading through joint segmentation~\cite{zhou2019collaborative} and learned lesion queries~\cite{sun2021lat}, usually within one network.

Retinal segmenters build on U-Net~\cite{ronneberger2015unet} with attention gates~\cite{oktay2018attention}, the convolutional block attention module (CBAM)~\cite{woo2018cbam}, the focal Tversky loss~\cite{abraham2019focaltversky} and, for vessels, clDice~\cite{shit2021cldice}. Sparse mixture-of-experts (MoE) layers with noisy top-$k$ gating~\cite{shazeer2017moe} add capacity at almost constant compute, but their experts are copies of one block~\cite{fedus2022switch}.

\subsection{Research Gaps and Contributions}
These studies leave three gaps that matter for a staged system. First, the four lesions differ widely in size, contrast and shape, yet segmenters typically apply the same blocks to all of them, and sparse MoE models replicate one block as their experts. Second, ungradable images are usually excluded or added as an extra class, and lesion-aware graders rarely test whether the trained grade changes when the lesion input is removed. Third, staged systems often report the grader only on the images that reached it, so errors passed from one stage to the next are not measured. \emph{HierarchiRetina} (\figref{fig:overview}) follows the clinical order and addresses each gap. Its contributions are:
\begin{enumerate}
\item \textbf{HSMoE-AUNet and SwinHRUNetPP (Stage~II).} HSMoE-AUNet (hierarchical sparse mixture-of-experts attention U-Net) is a U-Net whose skip processors, bottleneck and decoder blocks are sparse MoE blocks with experts of different \emph{types}, from dilated and large-kernel convolutions to top-hat and texture experts, so the router selects a processing strategy for each image. Each lesion has its own expert pool, expert counts and loss. SwinHRUNetPP fuses a SwinV2 encoder with a high-resolution convolutional neural network (CNN) branch under a field-of-view-restricted loss.
\item \textbf{LG-DRG (Stage~III).} The lesion-guided DR grader has separate gradability and rank-consistent ordinal heads, decision thresholds fitted on out-of-fold predictions only, and a zero-initialised cross-attention gate through which the masks query the image. A mask ablation, repeated with the gate started open, tests whether the masks drive the grade.
\item \textbf{End-to-end evaluation with error attribution.} Every screening miss is a final error, every error is assigned to its stage, and the DDR comparison uses matched protocols. The screening gate causes two thirds of all errors.
\end{enumerate}

\section{Materials and Methods}
\label{sec:methods}

\subsection{Data}
\label{sec:data}
\begin{figure*}[!t]
\centering
\includegraphics[width=0.86\textwidth]{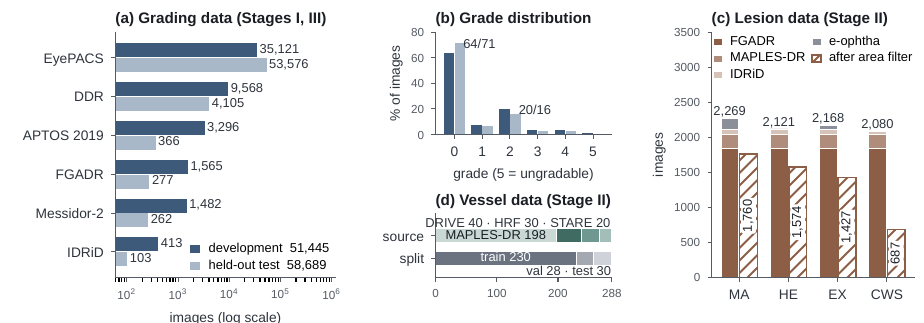}
\caption{Data. (a) Development and test images per grading source (log scale). (b) Grade distribution, development (dark) and test (light). (c) Lesion pools before (solid) and after (hatched) the area filter that removes near-empty masks. (d) Vessel pool and split.}
\label{fig:data}
\end{figure*}

\emph{Grading.} We pooled six public datasets (\figref{fig:data}), with Grades~0 to 4 for no DR to proliferative DR and Grade~5 for ungradable. EyePACS~\cite{cuadros2009eyepacs}, DDR~\cite{li2019ddr} and IDRiD~\cite{porwal2018idrid} kept their official test partitions. APTOS 2019~\cite{aptos2019}, which has no public labelled test set, was split 80/10/10 at random, and Messidor-2~\cite{decenciere2014messidor,abramoff2016improved} and FGADR~\cite{zhou2021fgadr} were each split once, 85/15. The held-out partitions form the test pool (58,689 images), never used for training or for choosing a model or threshold; the remaining 51,445 images form the development pool. The evaluation is therefore held-out rather than external.

\emph{Lesions and vessels.} Lesion image and mask pairs came from FGADR, MAPLES-DR~\cite{lepetitaimon2024maples}, IDRiD and e-ophtha~\cite{decenciere2013teleophta}. After removing near-empty masks (at most 10, 50, 30 and 20 positive pixels for MA, HE, EX and CWS), each pool was split 70/20/10. Vessels came from DRIVE~\cite{staal2004drive}, STARE~\cite{hoover2000stare}, HRF~\cite{budai2013hrf} and MAPLES-DR (288 images, split 230/28/30). Because the lesion pools were split independently of the grading partitions, 259 grading-test images were in a lesion training split and 27 in the vessel pool; excluding the affected sources (FGADR, Messidor-2 and, as it cannot be checked by name, IDRiD) changes the headline quadratic weighted kappa (QWK) only from 0.804 to 0.800 (\secref{sec:res-cascade}).

\subsection{Stage I: Screening Gate}
\label{sec:stage1}
Stage~I separates Grade~0 from Grades~1 to 5 (here, DR), keeping ungradable images positive so that they reach the gradability head. It uses standard components: a square retinal crop with contrast enhancement, a ConvNeXt~V2-L~\cite{woo2023convnextv2} with generalised-mean and attention pooling (201.7\,M parameters), a combined binary cross-entropy (BCE) and focal loss~\cite{lin2017focal} with Grade~1 over-sampled, exponential-moving-average (EMA) weights and four-view test-time augmentation. Resolution is set by the smallest lesion: microaneurysms, the only lesion that defines mild DR, span 2 to 3 pixels at $512^2$, so the gate works at $768^2$ without augmentations that erase small dots. Its threshold, $\tau=0.2392$, maximises $2\,\mathrm{TPR}-\mathrm{FPR}$ (true and false positive rates) on validation data.

\subsection{Stage II: HSMoE-AUNet for Lesions}
\label{sec:stage2}
\begin{figure*}[!t]
\centering
\includegraphics[width=\textwidth]{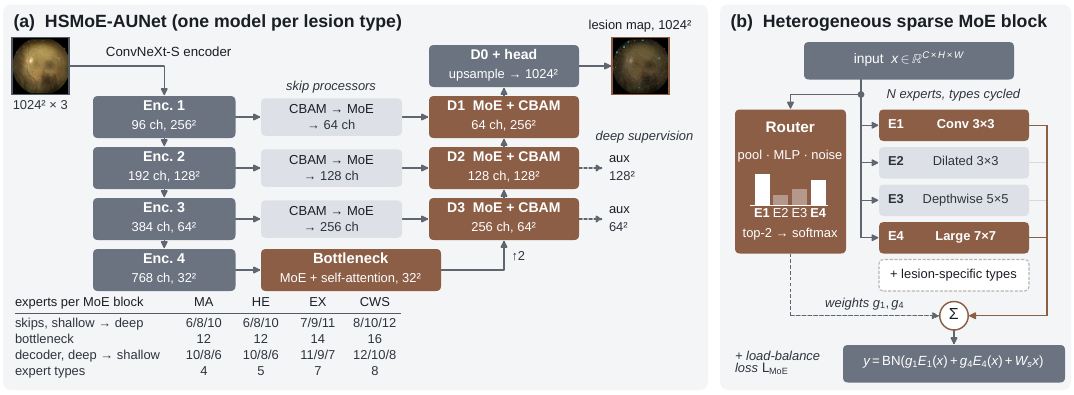}
\caption{HSMoE-AUNet. (a) ConvNeXt-S encoder, skip processors (CBAM followed by MoE), MoE bottleneck with self-attention, MoE decoder with CBAM and deep supervision; the inset lists expert counts per lesion. (b) The router scores $N$ heterogeneous experts and mixes the best two. MLP, multilayer perceptron.}
\label{fig:hsmoe}
\end{figure*}

At $1024^2$, MA are dark dots about 4 to 6 pixels across, HE larger dark blobs, EX bright and sharp-edged, and CWS pale and diffuse, so a single receptive field is unlikely to suit all four. HSMoE-AUNet (\figref{fig:hsmoe}) therefore makes every skip processor, the bottleneck and every decoder block a sparse MoE block, with one model per lesion.

\emph{Heterogeneous MoE block.} A router pools the block input $x$, scores $N$ experts with a two-layer perceptron ($C\to C/4\to N$, Gaussian noise $\sigma=0.01$ in training), keeps the two best and normalises their scores $g_i$ with a softmax~\cite{shazeer2017moe}:
\begin{equation}
\mathbf{y}=\mathrm{BN}\Big(\textstyle\sum_{i\in\mathcal{T}(x)} g_i(x)\,E_i(x)+W_s x\Big),
\label{eq:moe}
\end{equation}
where $\mathcal{T}(x)$ holds the two selected experts $E_i$, $W_s$ is a $1\times1$ projection and BN is batch normalisation. All experts are evaluated; only the mixture is sparse. The base pool has a $3\times3$ residual convolution for dots and edges, a dilated convolution for the lesion's surroundings, a depthwise-separable $5\times5$ convolution for texture and a $7\times7$ large-kernel convolution for large regions. Each lesion model adds the types its target needs (\tabref{tab:hsmoe}): a multi-scale expert for HE, EX and CWS, a morphological top-hat expert for the bright EX and CWS, a difference-of-Gaussians edge expert for the sharp-edged EX, and soft-region and texture experts for the diffuse CWS. A load-balancing loss $N\sum_i\bar p_i f_i$ (weight 0.01 per block), with $\bar p_i$ the mean routing probability of expert $i$ and $f_i$ the fraction of images routed to it, keeps all experts in use. \emph{Scale-adaptive capacity:} deeper blocks and rarer lesions receive more experts.

\emph{Architecture and loss.} A ConvNeXt-S encoder~\cite{liu2022convnext} feeds three skip processors (CBAM followed by an MoE block), an MoE bottleneck with 8-head self-attention over $32\times32$ tokens, and three MoE decoder blocks with CBAM; auxiliary heads at $64^2$ and $128^2$ add deep supervision. Because weight initialisation ran after the ImageNet weights were loaded, it reset the encoder's convolution and linear layers; the encoder was in effect trained from scratch, with its stem and first two blocks frozen at initialisation. Each model is trained on whole $1024^2$ images with focal Tversky~\cite{abraham2019focaltversky}, Dice, BCE and auxiliary losses, plus lesion-specific terms (\tabref{tab:hsmoe}): boundary terms, optic-disc suppression for EX and CWS, and region coherence for CWS. For CWS, the rarest lesion, these terms are phased in so that they do not dominate before any positives are predicted. On validation data, thresholds of 0.20 (HE) and 0.325 (EX) maximised Dice. For MA, Dice and recall kept rising as the threshold fell, and we used 0.01; for CWS we used 0.40, below the Dice-optimal 0.575, to favour recall.

\begin{table}[!t]
\centering
\caption{HSMoE-AUNet per lesion. Experts: skips (shallow to deep) / bottleneck / decoder (deep to shallow). Added types: MS multi-scale, TH top-hat, DoG difference-of-Gaussians, SR soft region, TX texture. FTL: focal Tversky loss (false-negative weight $a$, false-positive weight $b$, exponent). Extra losses: B boundary, OD optic disc, RC region coherence. LR: learning rate.}
\label{tab:hsmoe}
\footnotesize
\setlength{\tabcolsep}{2.0pt}
\begin{tabular}{@{}lcccc@{}}
\toprule
 & MA & HE & EX & CWS \\
\midrule
Expert types     & 4 & 5 & 7 & 8 \\
Added types      & -- & MS & MS, TH, DoG & MS, TH, SR, TX \\
Experts (skips)  & 6/8/10 & 6/8/10 & 7/9/11 & 8/10/12 \\
Experts (bottl.) & 12 & 12 & 14 & 16 \\
Experts (dec.)   & 10/8/6 & 10/8/6 & 11/9/7 & 12/10/8 \\
Parameters (M)   & 218.3 & 190.5 & 187.8 & 192.7 \\
FTL $a/b$/exp. & 0.30/0.70/2 & 0.50/0.50/$\frac43$ & 0.45/0.55/1 & 0.55/0.45/$\frac43$ \\
Extra loss       & -- & B & B, OD & B, RC, OD \\
LR schedule      & plateau & cos.\,restart & cosine & cos.\,restart \\
Best epoch       & 262 & 128 & 199 & 84 \\
Threshold        & 0.01 & 0.20 & 0.325 & 0.40 \\
\bottomrule
\end{tabular}
\end{table}

\subsection{Stage II: SwinHRUNetPP for Vessels}
\begin{figure*}[!t]
\centering
\includegraphics[width=\textwidth]{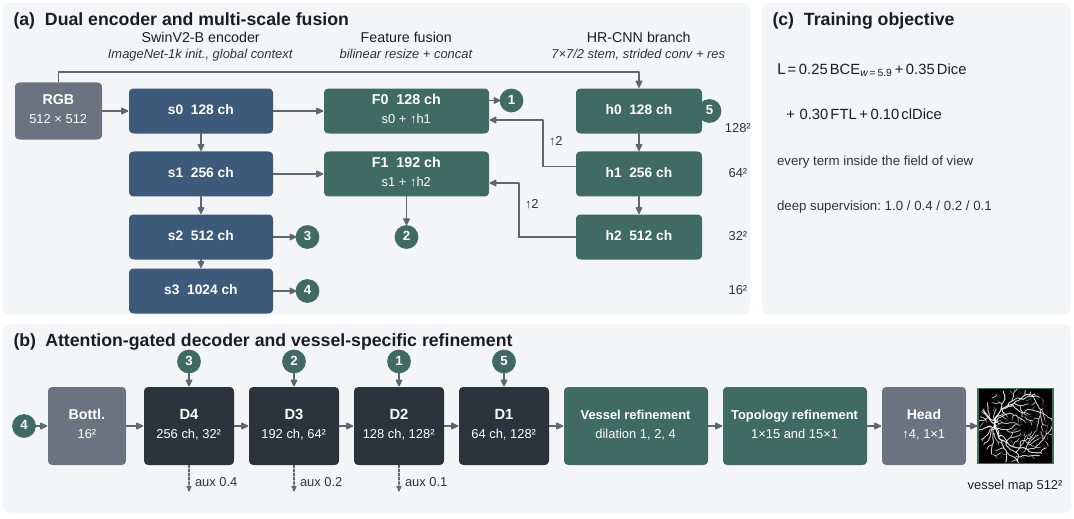}
\caption{SwinHRUNetPP. (a) SwinV2-B encoder and high-resolution CNN branch, fused at $128^2$ and $64^2$; numbered circles connect (a) to (b). (b) Bottleneck, attention-gated decoder, refinement and topology blocks; dashed arrows are auxiliary heads. (c) All loss terms are computed inside the FOV.}
\label{fig:swin}
\end{figure*}

SwinHRUNetPP (\figref{fig:swin}) runs a SwinV2-B encoder~\cite{liu2022swinv2}, whose shifted-window attention relates distant parts of the vessel tree, beside a high-resolution CNN branch that keeps capillary detail. Two fusion modules merge the streams at $128^2$ and $64^2$, and four attention-gated decoder blocks~\cite{oktay2018attention} follow. A refinement block then fuses parallel convolutions with dilations 1, 2 and 4 by squeeze-and-excitation~\cite{hu2018senet}, and a topology block applies elongated $1\times15$ and $15\times1$ kernels that match locally straight vessel segments. Without a field-of-view (FOV) mask, the bright fundus rim is traced as a vessel, so every loss term is computed inside the FOV: 0.25 BCE (positive weight 5.9), 0.35 Dice, 0.30 focal Tversky and 0.10 clDice~\cite{shit2021cldice}, with deep supervision on three decoder levels. The model (110.2\,M parameters, $512^2$ input) is thresholded at 0.46, the Youden-optimal validation value.

\subsection{Stage III: Lesion-Guided DR Grader (LG-DRG)}
\label{sec:stage3}
\begin{figure*}[!t]
\centering
\includegraphics[width=\textwidth]{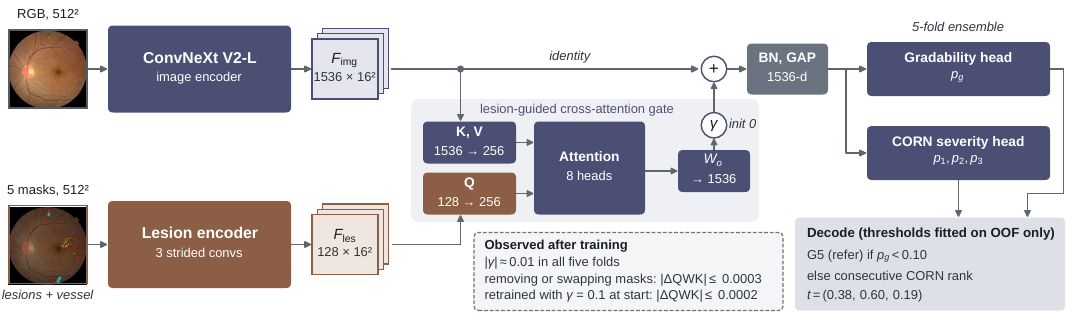}
\caption{LG-DRG. RGB and mask streams meet in a residual cross-attention gate scaled by $\gamma$ (initialised to 0); the pooled vector feeds a gradability head and a CORN severity head with OOF-fitted thresholds. The inset summarises \protect\secref{sec:ablation}. BN, batch normalisation; GAP, global average pooling.}
\label{fig:lgdrg}
\end{figure*}

LG-DRG (\figref{fig:lgdrg}) first decides whether an image can be graded and then estimates its severity.

\emph{Gated cross-attention.} A ConvNeXt~V2-L encodes the RGB image (fundus crop, contrast-limited adaptive histogram equalisation on LAB lightness, $512^2$) into $\mathbf{F}_{\mathrm{img}}\in\mathbb{R}^{1536\times16\times16}$, and a small CNN trained from scratch encodes the five masks into $\mathbf{F}_{\mathrm{les}}\in\mathbb{R}^{128\times16\times16}$. The lesion tokens query the image tokens through multi-head attention (MHA):
\begin{equation}
\mathbf{F}_{\mathrm{gated}}=\mathbf{F}_{\mathrm{img}}+\gamma\,W_o\,\mathrm{MHA}\big(W_q\mathbf{F}_{\mathrm{les}},W_k\mathbf{F}_{\mathrm{img}},W_v\mathbf{F}_{\mathrm{img}}\big),
\label{eq:gate}
\end{equation}
with 8 heads and a learned scalar $\gamma$ that starts at~0. The network thus starts as an image classifier, and the masks act only if training opens the gate, a safeguard against imperfect masks.

\emph{Two heads.} Ungradability reflects image quality rather than disease severity; placed on the ordinal axis, it would rank above proliferative DR. A gradability head, whose output $p_g$ estimates P(gradable), is trained on all images, and a conditional ordinal regression for neural networks (CORN) head~\cite{shi2023corn} models $p_k=P(y>k\mid y>k-1)$ over Grades~1 to 4 on gradable images. (In our implementation the second and third tasks were also trained on images one grade below the conditioning boundary, so $p_2$ and $p_3$ are not exact conditionals.) The output is Grade~5 if $p_g<t_g$ and otherwise the rank-consistent grade $1+\sum_{k}\prod_{j\le k}\mathbf{1}[p_j>t_j]$. All four thresholds were fitted by coordinate ascent on \emph{out-of-fold} (OOF) predictions: $t=(0.38,0.60,0.19)$ and $t_g=0.10$.

\emph{Training.} Five stratified fold models were trained on the 18,570 development images of Grades~1 to 5 with AdamW~\cite{loshchilov2019adamw} (weight decay 0.05, effective batch 64), EMA weights and inverse-frequency sampling, and selected on the macro-averaged area under the receiver operating characteristic curve (ROC AUC). At test time the fold-averaged probabilities are mapped to $(p_1,p_2,p_3,p_g)$ and decoded.

\subsection{Evaluation Protocol}
\label{sec:protocol}
\emph{View~B} scores all 58,689 test images end-to-end. An image stopped at Stage~I receives Grade~0, so a screening miss is a final error, and a healthy image that passes is always mis-graded (LG-DRG lacks Grade~0). \emph{View~A} scores the grader alone, on images of true Grades~1 to 5 that reached it. For QWK, true Grade~5 images are dropped and predicted Grade~5 counts as Grade~4; placing Grade~5 on the ordinal axis would raise QWK to 0.814. Referral means Grade~2 or higher or ungradable; vision-threatening DR means true Grade~3 or 4, detected if graded 3 or 4. The mask ablation re-scores the 3,714 fold-0 validation images without masks, with swapped masks and with each channel removed in turn. Confidence intervals (CIs) use 1,000 bootstrap resamples (500 for AUC).

\section{Results}
\label{sec:results}
\subsection{Stage I}
\begin{figure}[!t]
\centering
\includegraphics[width=\columnwidth]{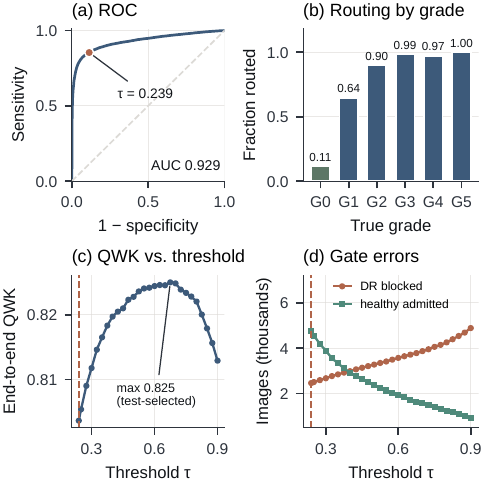}
\caption{Stage~I on the test pool. (a) Receiver operating characteristic (ROC) curve with the validation-derived operating point $\tau$. (b) Fraction of each true grade routed onwards. (c) End-to-end QWK against the Stage~I threshold; the maximum is test-selected and shown for illustration only. (d) DR cases blocked and healthy eyes admitted at each threshold. Dashed line: $\tau$.}
\label{fig:stage1}
\end{figure}

The $768^2$ gate reached AUC 0.929 (95\% CI 0.926 to 0.932) and, at an equal load of 19,154 routed images (\tabref{tab:stage1}), routed 64.5\% of mild-DR images, against 0.904 and 52.8\% for our earlier $512^2$ gate with dot-erasing augmentations, and 0.909 and 53.4\% for a seven-model ensemble. Higher resolution with dot-preserving augmentation helped more than ensembling. At $\tau$, sensitivity was 0.854 and specificity 0.887; the gate passed 97.8\% of severe and proliferative DR and 345 of 346 ungradable images, and stopped 67.4\% of the pool, including 2,459 DR cases, 97\% of them mild or moderate (\figref{fig:stage1}).

\begin{table*}[!t]
\centering
\caption{Stage~I on the 58,689-image test pool at an equal routing load of 19,154 images. G1 routed: fraction of mild-DR images routed onwards; FN: false negatives. $^{\dagger}$No leakage-free set exists to set this ensemble's threshold; reference only.}
\label{tab:stage1}
\footnotesize
\setlength{\tabcolsep}{6pt}
\begin{tabular}{@{}lcccccr@{}}
\toprule
Model & Input & AUC & Sens. & Spec. & G1 routed & FN \\
\midrule
SwinV2-L                         & 384 & 0.884 & 0.777 & 0.855 & 0.458 & 3,769 \\
ConvNeXt V2-L (fold models)      & 512 & 0.889--0.902 & 0.789--0.810 & 0.860--0.869 & 0.493--0.528 & 3,209--3,558 \\
ConvNeXt V2-L                    & 512 & 0.904 & 0.809 & 0.868 & 0.528 & 3,214 \\
Hybrid ensemble (7 models)       & mixed & 0.909 & 0.818 & 0.872 & 0.534 & 3,078 \\
\midrule
\tb{ConvNeXt V2-L (final)}        & \tb{768} & \tb{0.929} & \tb{0.854} & \tb{0.887} & \tb{0.645} & \tb{2,459} \\
\quad + hybrid, $5\times$ weight$^{\dagger}$ & mixed & 0.933 & 0.862 & 0.889 & 0.643 & 2,336 \\
\bottomrule
\end{tabular}
\end{table*}

\subsection{Stage II}
\label{sec:res-s2}
\begin{figure*}[!t]
\centering
\includegraphics[width=0.705\textwidth]{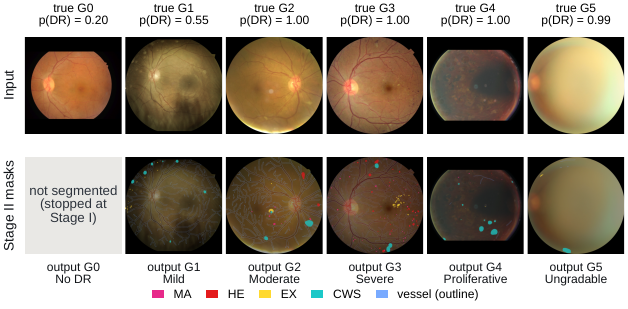}
\caption{One correctly graded test image per true grade: input with Stage~I probability (top) and Stage~II masks with the final output (bottom; magenta MA, red HE, yellow EX, cyan CWS, blue vessels).}
\label{fig:qual}
\end{figure*}

Exudates were segmented best (\tabref{tab:seg}; pooled Dice 0.579), haemorrhage masks were conservative (precision 0.685, sensitivity 0.407), and microaneurysms remained the hardest target (Dice 0.184). Every lesion model detected its lesion in 97\% to 100\% of the test images, all of which contain that lesion. Vessels reached pooled Dice 0.731 and ROC AUC 0.966. \figref{fig:qual} shows one test image of each grade.

\begin{table*}[!t]
\centering
\caption{Stage~II on held-out test splits (no lesion-free test images). Pixel metrics pooled over test pixels (inside the FOV for vessels); image Dice as mean $\pm$ SD; lesion level: connected components matched at intersection over union (IoU) $\geq0.3$ ($\geq0.2$ for MA); image sens.: fraction of images whose lesion was detected. PR, precision-recall.}
\label{tab:seg}
\footnotesize
\setlength{\tabcolsep}{3.2pt}
\begin{tabular}{@{}lrccccccccccc@{}}
\toprule
 & & & \multicolumn{6}{c}{Pixel, pooled} & Image & \multicolumn{2}{c}{Lesion level} & Image \\
\cmidrule(lr){4-9}\cmidrule(lr){11-12}
Target & $n$ & Thr. & Dice & IoU & Sens. & Prec. & ROC AUC & PR AUC & Dice & Sens. / Prec. & F1 & sens. \\
\midrule
Hard exudates     & 142 & 0.325 & 0.579 & 0.407 & 0.583 & 0.574 & 0.846 & 0.489 & $0.470\pm0.221$ & 0.476 / 0.473 & 0.475 & 1.000 \\
Haemorrhages      & 157 & 0.20  & 0.510 & 0.343 & 0.407 & 0.685 & 0.800 & 0.411 & $0.419\pm0.245$ & 0.412 / 0.445 & 0.428 & 1.000 \\
Cotton-wool spots &  69 & 0.40  & 0.408 & 0.257 & 0.521 & 0.336 & 0.884 & 0.328 & $0.353\pm0.261$ & 0.421 / 0.306 & 0.355 & 0.971 \\
Microaneurysms    & 175 & 0.01  & 0.184 & 0.101 & 0.137 & 0.277 & 0.749 & 0.072 & $0.242\pm0.169$ & 0.405 / 0.248 & 0.307 & 0.977 \\
\midrule
Vessels           &  30 & 0.46  & 0.731 & 0.576 & 0.915 & 0.609 & 0.966 & 0.852 & $0.727\pm0.043$ & -- & -- & -- \\
\bottomrule
\end{tabular}
\end{table*}

\subsection{Stage III and the Mask Ablation}
\label{sec:ablation}
With argmax decoding, OOF QWK was $0.771\pm0.011$ across folds (\tabref{tab:oof}). Ungradable images were separated almost perfectly (AUC 0.999), and the OOF-fitted thresholds raised pooled OOF QWK from 0.771 to 0.781.

\emph{Mask ablation.} In every fold $|\gamma|$ was about 0.01, and in fold~0 no ablation condition (\tabref{tab:ablation}) changed QWK by more than 0.0003: the cross-attention gate had stayed closed. Because weight decay could have held $\gamma$ near zero, we retrained fold~0 with $\gamma$ starting at 0.1 and free of weight decay. After 60 epochs $\gamma$ was still 0.100 (EMA weights), yet the grade remained insensitive to the masks ($|\Delta\mathrm{QWK}|\le0.0002$), and the model was slightly weaker (QWK 0.744 against 0.773). Weight decay therefore does not explain the result.

\begin{table}[!t]
\centering
\caption{Stage~III five-fold cross-validation ($n=18{,}570$). G5: Grade~5 (ungradable). Mean and SD from unrounded values.}
\label{tab:oof}
\footnotesize
\setlength{\tabcolsep}{3pt}
\begin{tabular}{@{}lccccc@{}}
\toprule
 & Macro-AUC & G5 AUC & QWK & Acc. & Macro-F1 \\
\midrule
Fold 0 & 0.942 & 0.999 & 0.771 & 0.758 & 0.767 \\
Fold 1 & 0.945 & 0.998 & 0.783 & 0.770 & 0.768 \\
Fold 2 & 0.938 & 0.997 & 0.754 & 0.745 & 0.751 \\
Fold 3 & 0.942 & 0.999 & 0.779 & 0.764 & 0.769 \\
Fold 4 & 0.939 & 0.999 & 0.769 & 0.756 & 0.763 \\
\midrule
Mean & 0.941 & 0.999 & 0.771 & 0.759 & 0.764 \\
SD   & 0.003 & 0.001 & 0.011 & 0.009 & 0.007 \\
Pooled, OOF-fitted & -- & -- & \textbf{0.781} & \textbf{0.774} & -- \\
\bottomrule
\end{tabular}
\end{table}

\begin{table}[!t]
\centering
\caption{Mask reliance of LG-DRG (fold~0, $n=3{,}714$). Original: $\gamma$ starts at 0, with weight decay. Retrained: $\gamma$ starts at 0.1, no weight decay, accumulation 8 (not 4). Default 0.5-threshold decode, hence the gap to \protect\tabref{tab:oof} (argmax); $\Delta$ from unrounded QWK.}
\label{tab:ablation}
\footnotesize
\setlength{\tabcolsep}{4pt}
\begin{tabular}{@{}lcccc@{}}
\toprule
 & \multicolumn{2}{c}{Original} & \multicolumn{2}{c}{Retrained} \\
\cmidrule(lr){2-3}\cmidrule(l){4-5}
Mask input & QWK & $\Delta$QWK & QWK & $\Delta$QWK \\
\midrule
Intact                   & 0.7726 & --        & 0.7444 & --        \\
All masks zeroed         & 0.7725 & $-0.0002$ & 0.7446 & $+0.0002$ \\
Masks from another image & 0.7724 & $-0.0003$ & 0.7444 & $0.0000$  \\
\midrule
Without MA               & 0.7726 & $0.0000$  & 0.7444 & $0.0000$  \\
Without HE               & 0.7726 & $0.0000$  & 0.7444 & $0.0000$  \\
Without EX               & 0.7726 & $0.0000$  & 0.7444 & $0.0000$  \\
Without CWS              & 0.7725 & $-0.0001$ & 0.7444 & $0.0000$  \\
Without vessel           & 0.7725 & $-0.0002$ & 0.7446 & $+0.0002$ \\
\midrule
Macro-AUC, intact        & \multicolumn{2}{c}{0.9422} & \multicolumn{2}{c}{0.9359} \\
Learned $\gamma$         & \multicolumn{2}{c}{$|\gamma|\approx0.010$ (all folds)} & \multicolumn{2}{c}{0.100 (EMA)} \\
\bottomrule
\end{tabular}
\end{table}

\begin{table}[!t]
\centering
\caption{Cascade on the test pool. View~B: all images. View~A: grader only, on images of true Grades~1 to 5 that reached it.}
\label{tab:cascade}
\footnotesize
\setlength{\tabcolsep}{3pt}
\begin{tabular}{@{}lrccc@{}}
\toprule
View & $n$ & QWK (95\% CI) & Acc. & Macro-F1 \\
\midrule
B, OOF-fitted & 58,689 & \tb{0.804} (0.798--0.809) & 0.814 & 0.659 \\
B, argmax  & 58,689 & 0.801 & -- & -- \\
A, OOF-fitted & 14,409 & 0.732 (0.722--0.743) & 0.743 & -- \\
A, argmax  & 14,409 & 0.721 & -- & -- \\
\midrule
\multicolumn{5}{@{}l}{\emph{EyePACS, DDR and APTOS 2019 only (no possible Stage~II overlap)}} \\
B, OOF-fitted & 58,047 & 0.800 & 0.815 & -- \\
A, OOF-fitted & 13,992 & 0.727 & -- & -- \\
\bottomrule
\end{tabular}
\end{table}

\begin{table}[!t]
\centering
\caption{DDR official test split, matched protocols. Published values as reported; n/r: not reported; $^{\ast}$ungradable as a sixth ordinal level.}
\label{tab:ddr}
\footnotesize
\setlength{\tabcolsep}{1.7pt}
\begin{tabular}{@{}llcc@{}}
\toprule
Method & Protocol & Acc. (\%) & QWK \\
\midrule
FA+KC-Net+R2~\cite{tian2023fakcnet} & 6-class, grader & 83.78 & n/r \\
Ours & 6-class, end-to-end & \tb{85.63} & 0.913$^{\ast}$ \\
\midrule
FA+KC-Net+R2~\cite{tian2023fakcnet} & 5-class, grader & 84.49 & 0.862 \\
RadFuse~\cite{mohsen2025radfuse} & 5-class, grader & 83.32 & 0.855 \\
Dual-SwinOrd~\cite{yu2026dualswinord} & 5-class, grader & 86.54 & \tb{0.904} \\
\midrule
Ours, ungradable $\to$ Grade~4 & 5-class, end-to-end & 85.10 & 0.878 \\
Ours, ungradable dropped ($n{=}3{,}678$) & 5-class, end-to-end & -- & 0.892 \\
Ours, ungradable $\to$ worst case & 5-class, end-to-end & 84.70 & 0.851 \\
\bottomrule
\end{tabular}
\end{table}

\begin{figure*}[!t]
\centering
\includegraphics[width=0.84\textwidth]{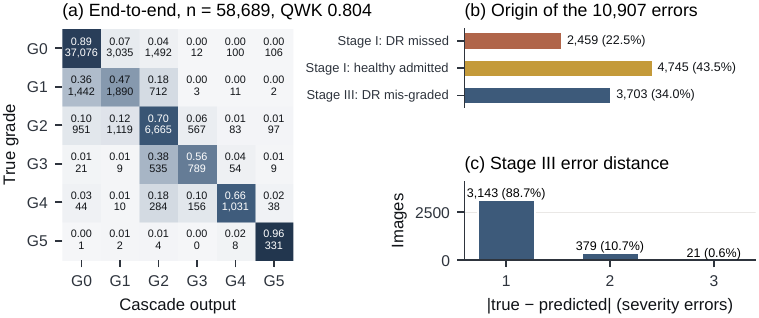}
\caption{Cascade errors. (a) End-to-end confusion matrix (row-normalised, with counts); (b) attribution of the 10,907 errors to stages; (c) distance of the 3,543 severity errors from the reference.}
\label{fig:cascade}
\end{figure*}

\begin{table*}[!t]
\centering
\caption{Per-dataset results. Gate sensitivity and specificity: Grade~0 versus Grades~1 to 5; A, B: View~A, View~B.}
\label{tab:perds}
\footnotesize
\setlength{\tabcolsep}{5pt}
\begin{tabular}{@{}lrrrcccccc@{}}
\toprule
Dataset & $n$ & DR & Grade 0 & Gate sens. & Gate spec. & DR blocked & QWK A & QWK B & Acc. B \\
\midrule
EyePACS    & 53,576 & 14,043 & 39,533 & 0.837 & 0.883 & 2,288 & 0.721 & 0.785 & 0.812 \\
DDR        &  4,105 &  2,225 &  1,880 & 0.931 & 0.969 &   154 & 0.750 & 0.878 & 0.856 \\
APTOS 2019 &    366 &    167 &    199 & 0.994 & 0.945 &     1 & 0.745 & 0.925 & 0.847 \\
FGADR      &    277 &    256 &     21 & 0.996 & 0/21  &     1 & 0.872 & 0.810 & 0.776 \\
Messidor-2 &    262 &    108 &    154 & 0.889 & 0.818 &    12 & 0.782 & 0.843 & 0.752 \\
IDRiD      &    103 &     69 &     34 & 0.957 & 0.647 &     3 & 0.747 & 0.830 & 0.660 \\
\midrule
All        & 58,689 & 16,868 & 41,821 & 0.854 & 0.887 & 2,459 & 0.732 & 0.804 & 0.814 \\
\bottomrule
\end{tabular}
\end{table*}

\subsection{End-to-End Cascade}
\label{sec:res-cascade}
End-to-end, the cascade reached QWK \tb{0.804} (95\% CI 0.798 to 0.809) and six-class accuracy 0.814 (\tabref{tab:cascade}). For referral, sensitivity was 0.832 and specificity 0.947. The grader alone (View~A) reached QWK 0.732, and the OOF-fitted thresholds improved on argmax decoding in both views.

\emph{Error attribution.} Of 10,907 end-to-end errors, 22.5\% were DR cases blocked at the screening gate and 43.5\% were healthy eyes that passed it. Only 34.0\% were grading errors (\figref[(b)]{fig:cascade}). Of the 3,543 severity errors on Grades~1 to 4, 88.7\% were off by one grade (\figref[(c)]{fig:cascade}). Sensitivity for vision-threatening DR was 0.681; 819 of the 950 misses were graded moderate.

\subsection{Effect of the Gate Threshold}
QWK can be recomputed for any gate threshold without new inference. It peaks at 0.825 near 0.675 (\figref[(c)]{fig:stage1}), while blocked DR cases rise from 2,459 to 3,779 (\figref[(d)]{fig:stage1}), 94\% of the 1,320 extra cases being mild or moderate. QWK rises because every admitted healthy eye is mis-graded, 36\% by two or more grades. At 0.675, referral sensitivity would fall from 0.832 to 0.795 and specificity rise from 0.947 to 0.969.

\subsection{Protocol-Matched Comparison on DDR}
\label{sec:ddr}
Published DDR results use six classes on all 4,105 test images or five classes on the 3,759 gradable ones (\tabref{tab:ddr}). On the six-class protocol the cascade reached 85.6\% accuracy (95\% CI 84.6 to 86.7), above the 83.8\% of FA+KC-Net+R2~\cite{tian2023fakcnet}, the best six-class result we found, even though ours is end-to-end; recall was higher in four of the six classes, including Grade~3 (52.1\% vs.\ 39.4\%) and Grade~4 (82.2\% vs.\ 67.3\%). With our 81 ungradable calls on gradable images mapped to Grade~4, five-class QWK was 0.878, above FA+KC-Net+R2 (0.862) and RadFuse (0.855)~\cite{mohsen2025radfuse} but below Dual-SwinOrd (0.904)~\cite{yu2026dualswinord}; the worst-case mapping gave 0.851. Per-dataset end-to-end QWK ranged from 0.785 (EyePACS, 91\% of the pool) to 0.925 (APTOS 2019; \tabref{tab:perds}).

\section{Discussion}
\label{sec:discussion}
\emph{Lesion segmentation.} HSMoE-AUNet gave usable masks for EX, HE and CWS, and SwinHRUNetPP a usable vessel tree, but MA, only a few pixels across, remained weak (Dice 0.184). With the encoder trained from scratch and no identical-expert baseline, the benefit of heterogeneity itself is not yet isolated.

\emph{Gradability and lesion guidance.} The gradability head flagged 331 of the 346 ungradable test images (OOF AUC 0.999), with 252 false calls. The grade, however, ignored the masks even when the cross-attention gate stayed open, possibly because the RGB stream already carries the lesion evidence. Interpretability claims for mask-conditioned graders should thus be tested by removing or swapping masks; here the masks support review but do not explain the grade.

\emph{Cascade design.} With two thirds of the errors at the gate, a perfect grader would remove only a third. QWK thus favours a stricter gate than screening, which likely weighs the extra mild and moderate misses above admitted healthy eyes. We therefore kept the validation-derived $\tau$; end-to-end QWK should be reported with the gate's operating point and per-stage errors. Most missed vision-threatening cases were graded moderate, which keeps them referable but understates their urgency.

\emph{Comparison with published results.} The handling of our 81 ungradable calls alone moved five-class DDR QWK from 0.851 to 0.892, nearly the spread of published results (0.855 to 0.904), and our models were not trained on DDR alone. DDR results are therefore comparable only under a stated protocol.

\section{Limitations and Conclusions}
\label{sec:conclusion}
\emph{Limitations.} Test partitions share sources with the training data, EyePACS supplies 91\% of the test pool, Grade~5 occurs only in DDR, and most images were graded once, without adjudication~\cite{krause2018grader}. The segmenters were not compared with other architectures on the same splits.

\emph{Future directions.} The next steps are external, ideally prospective, validation with adjudicated grades~\cite{krause2018grader}; an identical-expert baseline with the pretrained encoder restored, and native-resolution patches for MA; gate and ordinal thresholds refitted, without retraining, to a screening cost that penalises missed referable disease and under-grading; and a re-test of lesion guidance on weaker image evidence, such as handheld-camera images.

\emph{Conclusions.} HierarchiRetina screens, segments and grades fundus images in explicit stages. It reached an end-to-end QWK of 0.804 on 58,689 held-out images and the highest six-class DDR accuracy we found. Most of its errors arose at the screening gate, and its grade did not use the lesion masks; staged systems should therefore be evaluated end-to-end, and their use of lesion inputs verified by ablation.

\section*{Data Availability and Ethics}
All datasets are public or available on request, and only de-identified images collected under the providers' approvals were used. Code, trained weights, per-image predictions and a script reproducing all end-to-end results are at \url{https://github.com/asifuddin01/HierarchiRetina}.

\bibliographystyle{IEEEtran}
\bibliography{references}

\begin{thebibliography}{10}
\providecommand{\url}[1]{#1}
\csname url@samestyle\endcsname
\providecommand{\newblock}{\relax}
\providecommand{\bibinfo}[2]{#2}
\providecommand{\BIBentrySTDinterwordspacing}{\spaceskip=0pt\relax}
\providecommand{\BIBentryALTinterwordstretchfactor}{4}
\providecommand{\BIBentryALTinterwordspacing}{\spaceskip=\fontdimen2\font plus
\BIBentryALTinterwordstretchfactor\fontdimen3\font minus
  \fontdimen4\font\relax}
\providecommand{\BIBforeignlanguage}[2]{{%
\expandafter\ifx\csname l@#1\endcsname\relax
\typeout{** WARNING: IEEEtran.bst: No hyphenation pattern has been}%
\typeout{** loaded for the language `#1'. Using the pattern for}%
\typeout{** the default language instead.}%
\else
\language=\csname l@#1\endcsname
\fi
#2}}
\providecommand{\BIBdecl}{\relax}
\BIBdecl

\bibitem{teo2021global}
Z.~L. Teo \emph{et~al.}, ``Global prevalence of diabetic retinopathy and
  projection of burden through 2045: Systematic review and meta-analysis,''
  \emph{Ophthalmology}, vol. 128, no.~11, pp. 1580--1591, 2021.

\bibitem{wilkinson2003proposed}
C.~P. Wilkinson \emph{et~al.}, ``Proposed international clinical diabetic
  retinopathy and diabetic macular edema disease severity scales,''
  \emph{Ophthalmology}, vol. 110, no.~9, pp. 1677--1682, 2003.

\bibitem{gulshan2016development}
V.~Gulshan \emph{et~al.}, ``Development and validation of a deep learning
  algorithm for detection of diabetic retinopathy in retinal fundus
  photographs,'' \emph{JAMA}, vol. 316, no.~22, pp. 2402--2410, 2016.

\bibitem{li2019ddr}
T.~Li \emph{et~al.}, ``Diagnostic assessment of deep learning algorithms for
  diabetic retinopathy screening,'' \emph{Inf. Sci.}, vol. 501, pp. 511--522,
  2019.

\bibitem{yu2026dualswinord}
W.~Yu, X.~Si, and J.~Zhong, ``{Dual-SwinOrd}: A dual-head {Swin} transformer
  with semantic prior injection for ordinal diabetic retinopathy grading,''
  \emph{Bioengineering}, vol.~13, no.~4, p. 374, 2026.

\bibitem{mohsen2025radfuse}
F.~Mohsen, S.~Belhaouari, and Z.~Shah, ``Integrating non-linear {R}adon
  transformation for diabetic retinopathy grading,'' \emph{Sci. Rep.}, vol.~15,
  p. 30706, 2025.

\bibitem{tian2023fakcnet}
M.~Tian \emph{et~al.}, ``Fine-grained attention \& knowledge-based
  collaborative network for diabetic retinopathy grading,'' \emph{Heliyon},
  vol.~9, no.~7, p. e17217, 2023.

\bibitem{zhou2019collaborative}
Y.~Zhou \emph{et~al.}, ``Collaborative learning of semi-supervised segmentation
  and classification for medical images,'' in \emph{Proc. IEEE/CVF CVPR}, 2019,
  pp. 2079--2088.

\bibitem{sun2021lat}
R.~Sun \emph{et~al.}, ``Lesion-aware transformers for diabetic retinopathy
  grading,'' in \emph{Proc. IEEE/CVF CVPR}, 2021, pp. 10\,938--10\,947.

\bibitem{ronneberger2015unet}
O.~Ronneberger, P.~Fischer, and T.~Brox, ``{U-Net}: Convolutional networks for
  biomedical image segmentation,'' in \emph{Proc. MICCAI}, 2015, pp. 234--241.

\bibitem{oktay2018attention}
O.~Oktay \emph{et~al.}, ``Attention {U-Net}: Learning where to look for the
  pancreas,'' in \emph{Proc. MIDL}, 2018.

\bibitem{woo2018cbam}
S.~Woo \emph{et~al.}, ``{CBAM}: Convolutional block attention module,'' in
  \emph{Proc. ECCV}, 2018, pp. 3--19.

\bibitem{abraham2019focaltversky}
N.~Abraham and N.~M. Khan, ``A novel focal {Tversky} loss function with
  improved attention {U-Net} for lesion segmentation,'' in \emph{Proc. IEEE
  ISBI}, 2019, pp. 683--687.

\bibitem{shit2021cldice}
S.~Shit \emph{et~al.}, ``{clDice}: A novel topology-preserving loss function
  for tubular structure segmentation,'' in \emph{Proc. IEEE/CVF CVPR}, 2021,
  pp. 16\,560--16\,569.

\bibitem{shazeer2017moe}
N.~Shazeer \emph{et~al.}, ``Outrageously large neural networks: The
  sparsely-gated mixture-of-experts layer,'' in \emph{Proc. ICLR}, 2017.

\bibitem{fedus2022switch}
W.~Fedus, B.~Zoph, and N.~Shazeer, ``Switch transformers: Scaling to trillion
  parameter models with simple and efficient sparsity,'' \emph{J. Mach. Learn.
  Res.}, vol.~23, no. 120, pp. 1--39, 2022.

\bibitem{cuadros2009eyepacs}
J.~Cuadros and G.~Bresnick, ``{EyePACS}: An adaptable telemedicine system for
  diabetic retinopathy screening,'' \emph{J. Diabetes Sci. Technol.}, vol.~3,
  no.~3, pp. 509--516, 2009.

\bibitem{porwal2018idrid}
P.~Porwal \emph{et~al.}, ``Indian diabetic retinopathy image dataset ({IDRiD}):
  A database for diabetic retinopathy screening research,'' \emph{Data},
  vol.~3, no.~3, p.~25, 2018.

\bibitem{aptos2019}
\BIBentryALTinterwordspacing
{Asia Pacific Tele-Ophthalmology Society}, ``{APTOS} 2019 blindness
  detection,'' Kaggle competition, 2019. [Online]. Available:
  \url{https://www.kaggle.com/c/aptos2019-blindness-detection}
\BIBentrySTDinterwordspacing

\bibitem{decenciere2014messidor}
E.~Decenci{\`e}re \emph{et~al.}, ``Feedback on a publicly distributed image
  database: The {Messidor} database,'' \emph{Image Anal. Stereol.}, vol.~33,
  no.~3, pp. 231--234, 2014.

\bibitem{abramoff2016improved}
M.~D. Abr{\`a}moff \emph{et~al.}, ``Improved automated detection of diabetic
  retinopathy on a publicly available dataset through integration of deep
  learning,'' \emph{Invest. Ophthalmol. Vis. Sci.}, vol.~57, no.~13, pp.
  5200--5206, 2016.

\bibitem{zhou2021fgadr}
Y.~Zhou \emph{et~al.}, ``A benchmark for studying diabetic retinopathy:
  Segmentation, grading, and transferability,'' \emph{IEEE Trans. Med. Imag.},
  vol.~40, no.~3, pp. 818--828, 2021.

\bibitem{lepetitaimon2024maples}
G.~Lepetit-Aimon \emph{et~al.}, ``{MAPLES-DR}: {MESSIDOR} anatomical and
  pathological labels for explainable screening of diabetic retinopathy,''
  \emph{Sci. Data}, vol.~11, no.~1, p. 914, 2024.

\bibitem{decenciere2013teleophta}
E.~Decenci{\`e}re \emph{et~al.}, ``{TeleOphta}: Machine learning and image
  processing methods for teleophthalmology,'' \emph{IRBM}, vol.~34, no.~2, pp.
  196--203, 2013.

\bibitem{staal2004drive}
J.~Staal \emph{et~al.}, ``Ridge-based vessel segmentation in color images of
  the retina,'' \emph{IEEE Trans. Med. Imag.}, vol.~23, no.~4, pp. 501--509,
  2004.

\bibitem{hoover2000stare}
A.~Hoover, V.~Kouznetsova, and M.~Goldbaum, ``Locating blood vessels in retinal
  images by piecewise threshold probing of a matched filter response,''
  \emph{IEEE Trans. Med. Imag.}, vol.~19, no.~3, pp. 203--210, 2000.

\bibitem{budai2013hrf}
A.~Budai \emph{et~al.}, ``Robust vessel segmentation in fundus images,''
  \emph{Int. J. Biomed. Imag.}, vol. 2013, p. 154860, 2013.

\bibitem{woo2023convnextv2}
S.~Woo \emph{et~al.}, ``{ConvNeXt V2}: Co-designing and scaling {ConvNets} with
  masked autoencoders,'' in \emph{Proc. IEEE/CVF CVPR}, 2023, pp.
  16\,133--16\,142.

\bibitem{lin2017focal}
T.-Y. Lin \emph{et~al.}, ``Focal loss for dense object detection,'' in
  \emph{Proc. IEEE ICCV}, 2017, pp. 2980--2988.

\bibitem{liu2022convnext}
Z.~Liu \emph{et~al.}, ``A {ConvNet} for the 2020s,'' in \emph{Proc. IEEE/CVF
  CVPR}, 2022, pp. 11\,976--11\,986.

\bibitem{liu2022swinv2}
Z.~Liu \emph{et~al.}, ``Swin transformer {V2}: Scaling up capacity and
  resolution,'' in \emph{Proc. IEEE/CVF CVPR}, 2022, pp. 12\,009--12\,019.

\bibitem{hu2018senet}
J.~Hu, L.~Shen, and G.~Sun, ``Squeeze-and-excitation networks,'' in \emph{Proc.
  IEEE/CVF CVPR}, 2018, pp. 7132--7141.

\bibitem{shi2023corn}
X.~Shi, W.~Cao, and S.~Raschka, ``Deep neural networks for rank-consistent
  ordinal regression based on conditional probabilities,'' \emph{Pattern Anal.
  Appl.}, vol.~26, no.~3, pp. 941--955, 2023.

\bibitem{loshchilov2019adamw}
I.~Loshchilov and F.~Hutter, ``Decoupled weight decay regularization,'' in
  \emph{Proc. ICLR}, 2019.

\bibitem{krause2018grader}
J.~Krause \emph{et~al.}, ``Grader variability and the importance of reference
  standards for evaluating machine learning models for diabetic retinopathy,''
  \emph{Ophthalmology}, vol. 125, no.~8, pp. 1264--1272, 2018.

\end{thebibliography}
\end{document}